\pdfoutput=1

\documentclass[11pt]{article}

\usepackage[preprint]{acl}

\usepackage{times}
\usepackage{latexsym}
\usepackage{booktabs}

\usepackage[T1]{fontenc}

\usepackage[utf8]{inputenc}

\usepackage{microtype}

\usepackage{inconsolata}

\usepackage{graphicx}
\usepackage{enumitem}
\usepackage{stfloats}
\usepackage{multirow}
\usepackage{makecell}
\usepackage{algorithm}
\usepackage[noend]{algorithmic}
\usepackage{amsmath}

\title{UGen: Unified Autoregressive Multimodal Model  with \\ Progressive Vocabulary Learning}

\author{
  \textbf{Hongxuan Tang\textsuperscript{1}}\textsuperscript{*},
  \textbf{Hao Liu\textsuperscript{1}}\textsuperscript{*},
  \textbf{Xinyan Xiao\textsuperscript{1}\thanks{\ \ Equal contribution.}},
\\
\\
  \textsuperscript{1}Baidu Inc., Beijing, China,
\\
 \{{\tt tanghongxuan,\ \ liuhao24,\ \ xiaoxinyan\}}@baidu.com
}

\begin{document}
\maketitle

\begin{abstract}
We introduce {\bf UGen}, a unified autoregressive multimodal model that demonstrates strong performance across text processing, image understanding, and image generation tasks simultaneously. UGen converts both texts and images into discrete token sequences and utilizes a single transformer to generate them uniformly in an autoregressive manner. To address the challenges associated with unified multimodal learning, UGen is trained using a novel mechanism, namely progressive vocabulary learning. In this process, visual token IDs are incrementally activated and integrated into the training phase, ultimately enhancing the effectiveness of unified multimodal learning. Experiments on comprehensive text and image tasks show that UGen achieves a significant overall performance improvement of 13.3\% compared to the vanilla unified autoregressive method, and it also delivers competitive results across all tasks against several task-specific models. 
\end{abstract}

\section{Introduction}
\label{Introduction}
Recently, researchers have sought to employ a single transformer to unify the language-vision understanding and generation, enabling the emergence of multimodal ability and the mutual enhancement of multimodal tasks. In most work, additional compositions are incorporated with the aim of achieving high performance across diverse multimodal tasks. To unify image understanding and generation, various approaches are proposed, including autoregressive models combined with diffusion-like methods \cite{zhou2024transfusion, xie2024show}, decoupling visual encoding approach \cite{wu2024janus}, and additional fine-tuning techniques \cite{wang2024emu3}. Furthermore, to maintain the language capability of the unified model, when is initialized from a pretrained language model, additional visual components and parameters are introduced \cite{shi2024llamafusion}. Although these methods differ, the introduction of additional compositions significantly increases the complexity of the unified model.

\begin{figure}[t]
\centering
\includegraphics[width=1.07\linewidth]{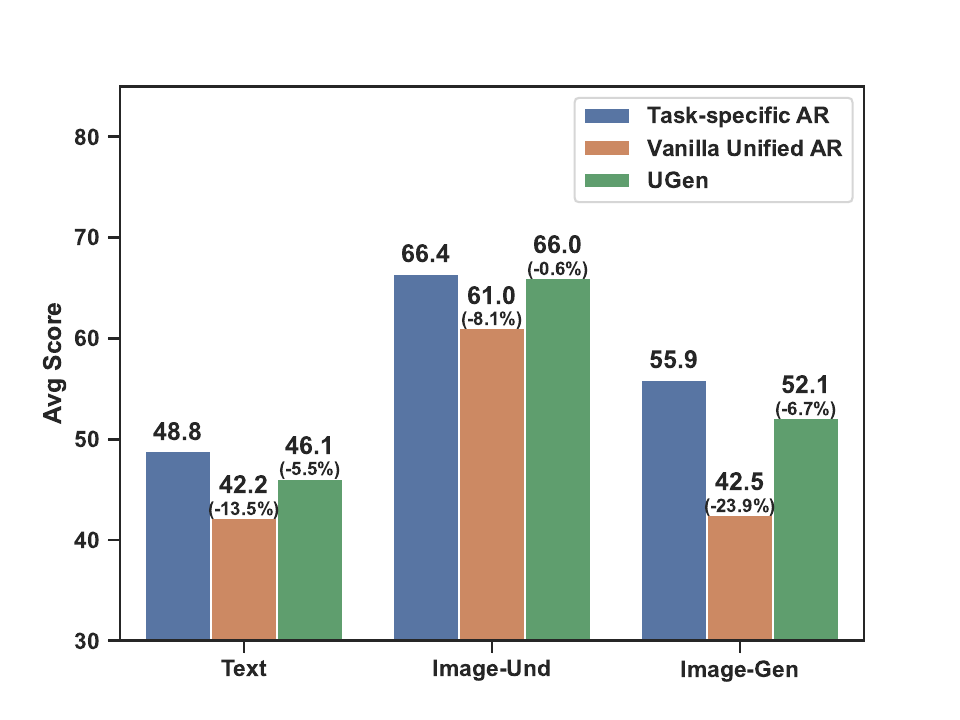}
\caption{The performance comparison among task-specific autoregressive models (Task-specific AR), current vanilla unified autoregressive model (Vanilla Unified AR) and UGen. Specifically, Text, Image-Und, Image-Gen denote text processing, image understanding and image generation tasks. Task-specific AR models are separated autoregressive models trained with single task data respectively. Vanilla Unified AR model is current unified autoregressive model trained with traditional joint learning approach \cite{wu2024liquid}.}

\label{spe}
\end{figure}

We believe a promising paradigm is the simple and direct unified autoregressive model, which only relies on a single unified transformer to generate discrete token sequences for both images and texts in an autoregressive manner, without any additional compositions. This approach enables to efficiently inherit the effectiveness and scalability of large language models (LLMs) \cite{touvron2023llama}. Furthermore, the fully unified transformer facilitates deeper fusion among modalities, thereby potentially achieving superior performance \cite{bubeck2023sparks}. Despite the great potential, current vanilla unified autoregressive models \cite{team2024chameleon,wu2024liquid} fail to achieve high performance on text processing, image understanding, and image generation tasks concurrently in practice. In our experiments, current vanilla unified autoregressive model exhibits an 8.1\% to 23.9\%  significant decrease in performance (Figure \ref{spe}), compared with the task-specific autoregressive method, which trains separated model for single task respectively.


\begin{figure*}[t]
\centering
\includegraphics[width=1\textwidth]{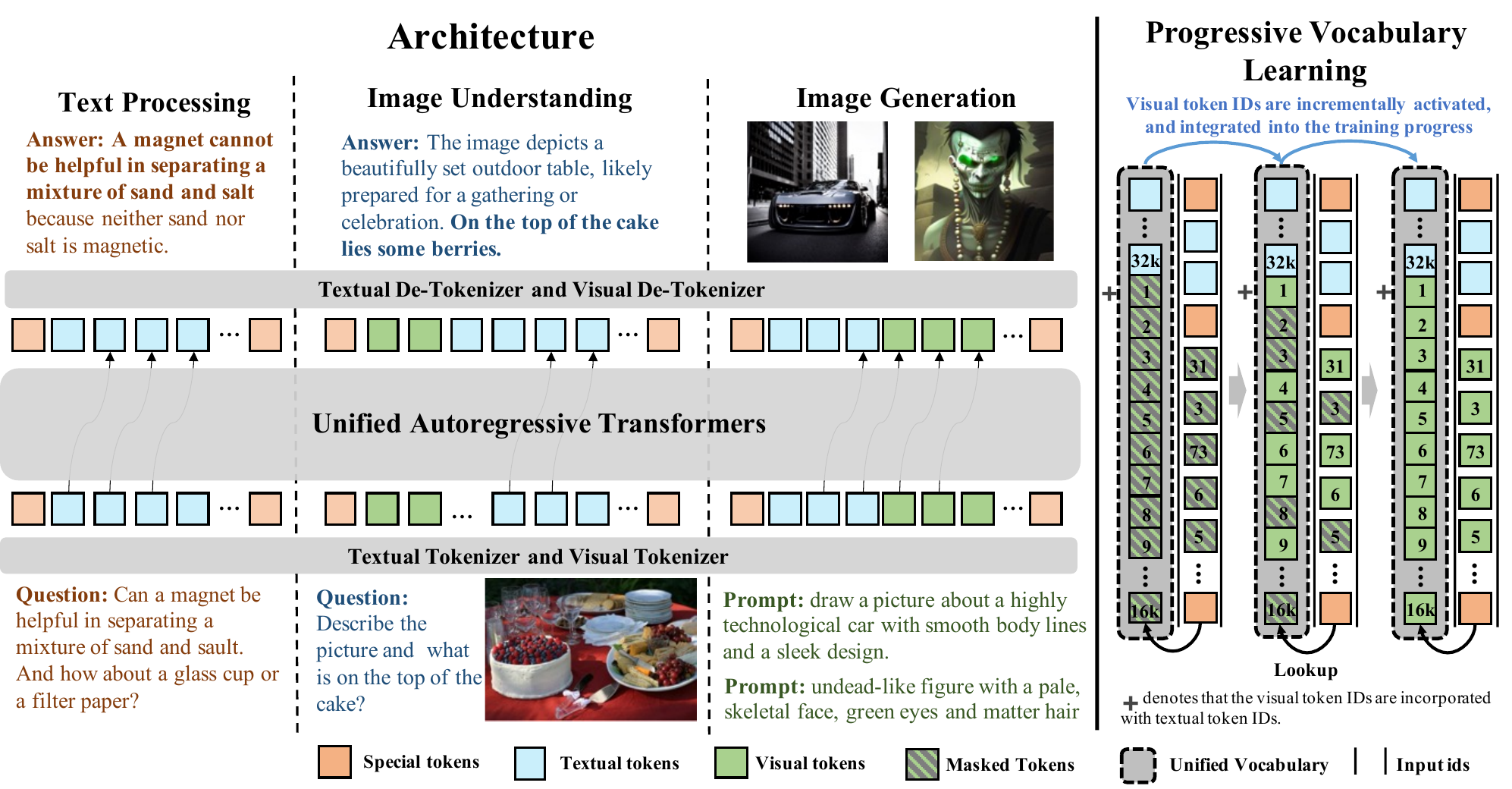}
\caption{An Overview of UGen. Left: A unified autoregressive 
generative architecture for language-vision understanding and generation. Both texts and images are converted into discrete token sequences and a single transformer is applied to generate them uniformly in an autoregressive manner. Right: The illustration of progressive vocabulary learning, which the visual token IDs are gradually activated and integrated into the training process.}
\label{arc}
\end{figure*}


In this paper, we aim to maintain the simple autoregressive model architecture, meanwhile achieving excellent performance in all language-vision understanding and generation tasks. To this end, we introduce UGen, which converts both texts and images into discrete token sequences and utilizes a single transformer to generate them uniformly in an autoregressive manner. To enhance the performance on various tasks, we propose a novel progressive vocabulary learning method. Specifically, during pretraining, the textual vocabulary is first learned separately, which enables the model to acquire strong text processing capability. Second, the visual token IDs are incrementally activated and integrated into the training process, which is designed to acquire the capabilities of image understanding and generation progressively. This approach is proved to be effective in improving the deep fusion among modalities and ultimately enhancing the model's performance on various tasks. Overall, our contributions are as follows:

\begin{itemize}[label=\textbullet, leftmargin=*]
    \item We introduce UGen, which employs an autoregressive generative architecture for unified multimodal understanding and generation. A novel progressive vocabulary learning method is proposed to enhance performance across text processing, image understanding and image generation tasks.  
    \item UGen significant outperforms the vanilla unified autoregressive  methods on diverse language-vision understanding and generation benchmarks, which strongly demonstrates the effectiveness of our proposed method. Furthermore, UGen also delivers competitive results against several individual task-specific models with an equivalent or larger number of parameters across various tasks. From the qualitative results of UGen, it also demonstrates the potential for mixed-modality generation.
    \item To the best of our knowledge, this is the first work to study and achieve strong performance on all the language-vision understanding and generation tasks only with a simple autoregressive architecture without any additional compositions. 
\end{itemize}

\section{UGen}

In this section, we firstly introduce the architecture of UGen (Section \ref{Architecture}), and then provide a detailed explanation of the progressive vocabulary learning method (Section \ref{Progressive Vocabulary Learning}) for performance enhancement. Finally, we present the training and inference procedure of UGen in Section \ref{Training and Inference Procedure of UGen}.



\subsection{Architecture}
\label{Architecture}
The architecture of UGen is shown in Figure \ref{arc}. UGen unifies language-vision understanding and generation through a single autoregressive transformer, eliminating the need for additional compositions required in most existing approaches \cite{xie2024show,wu2024janus}. The implementation details of UGen is illustrated as follows.

\paragraph{Tokenization} UGen uniformly converts both texts and images into discrete tokens, and then generation multimodal tokens in an autoregressive manner. Specifically, for text data, we employ the built-in BPE tokenizer \cite{sennrich2015neural} of LLMs to obtain discrete textual tokens. For image data, we select VQ-VAE \cite{van2017neural} to convert the image into discrete visual tokens. 

\begin{figure}[t]
\centering
\includegraphics[width=0.5\textwidth]{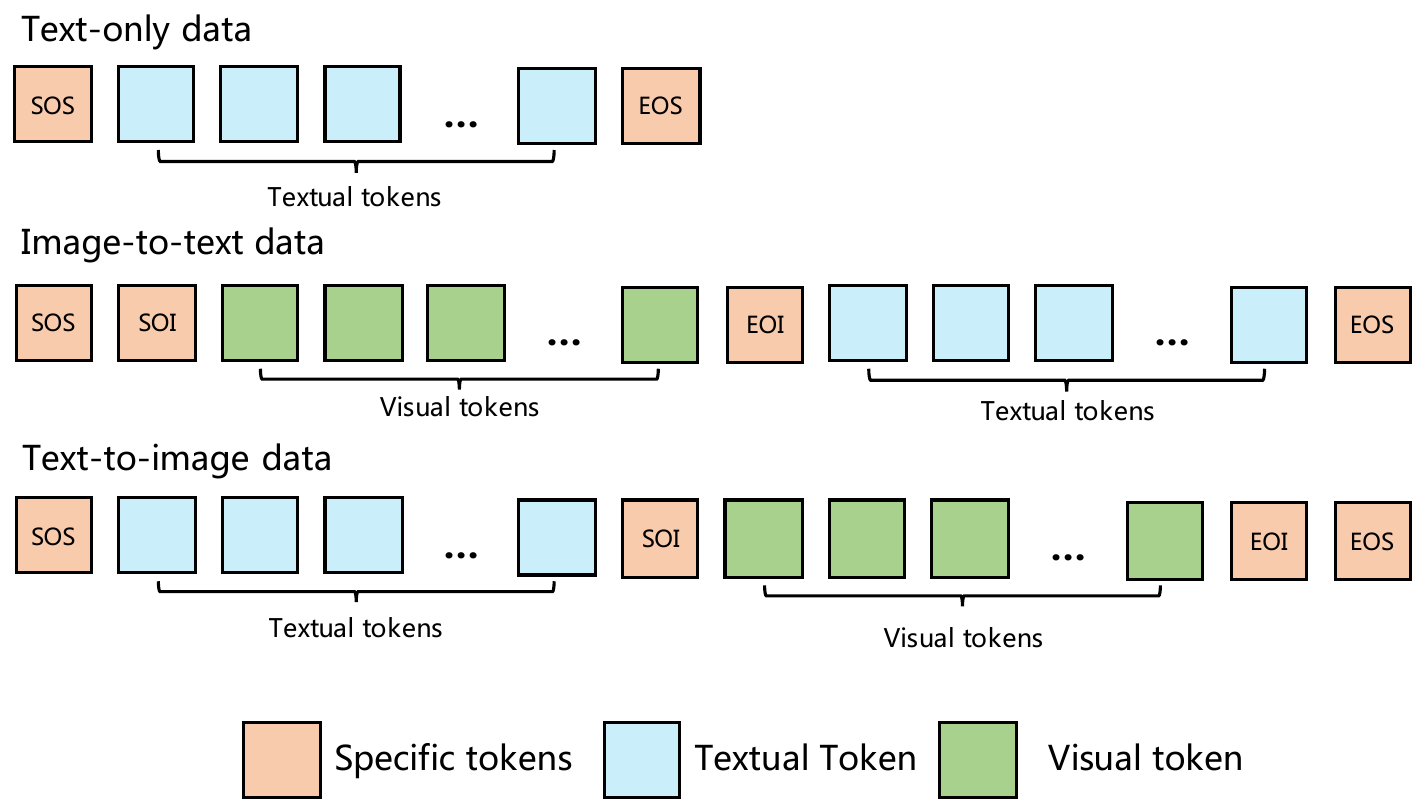}
\caption{Illustration of unified prompting.}
\label{data prompt}
\end{figure}

\paragraph{Unified Prompting} To perform unified learning on language-vision understanding and generation tasks, a unified prompting strategy is designed to format text-only, image-to-text and text-to-image input data. The detailed illustration of the unified prompting is shown in figure \ref{data prompt}. Specifically, $[SOI]$ and $[EOI]$ are pre-defined tokens denoting the start and end to image data. $[SOS]$ and $[EOS]$ serve as specical tokens to indicate the start and end of a training sample sequence. Given an text-image pair $(\textbf{x}, \textbf{y})$, it is first tokenized into $M$ textual tokens $\{x_1, x_2,...,x_M\}$ and $N$ visual tokens $\{y_1, y_2,..., y_N\}$ respectively. Then the text-only, image-to-text and text-to-image input data $\textbf{x}$, $(\textbf{y}, \textbf{x})$ and $(\textbf{x}, \textbf{y})$ can be formatted as follows.

\begin{equation}
\textbf{x} = \{[SOS], x_1, x_2, ..., x_M, [EOS]\}
\end{equation}

\begin{equation}
\begin{aligned}
& (\textbf{y}, \textbf{x}) = \{[SOS], [SOI], y_1, y_2, ..., y_N, \\ 
& [EOI], x_1, x_2,..., x_M, [EOS]\}
\end{aligned}
\end{equation}

\begin{equation}
\begin{aligned}
& (\textbf{x}, \textbf{y}) = \{[SOS], x_1, x_2, ..., x_M, \\ 
& [SOI], y_1, y_2,..., y_N, [EOI], [EOS]\}
\end{aligned}
\end{equation}



\paragraph{Autoregressive Transformer} By employing above unified prompting, we get formatted multimodal token sequences for various kinds of input data. Then the token sequences are fed into a single autoregressive transformer for processing. In detail, we select the same transformer architecture as TinyLlama \cite{zhang2024tinyllama}, which is widely used in open-source society \cite{zhao2024monoformer, zhou2024tinyllava}. TinyLlama is trained with various advanced techniques such as FlashAttention \cite{dao2023flashattention}, Lit-GPT \cite{lightning2023litgpt}. And it has demonstrated to achieve remarkable performance in a serious of downstream tasks and significantly outperform open-source LLMs with comparable sizes \cite{zhang2024tinyllama}. In addition, we obtain each textual token's feature representation from the original textual embeddings of TinyLlama. While the feature representations corresponding to each visual token are obtained from additional visual embeddings, which are initialized randomly and updated during training. 


\subsection{Progressive Vocabulary Learning}
\label{Progressive Vocabulary Learning}

\begin{figure}[t]
\centering
\includegraphics[width=1.0\linewidth]{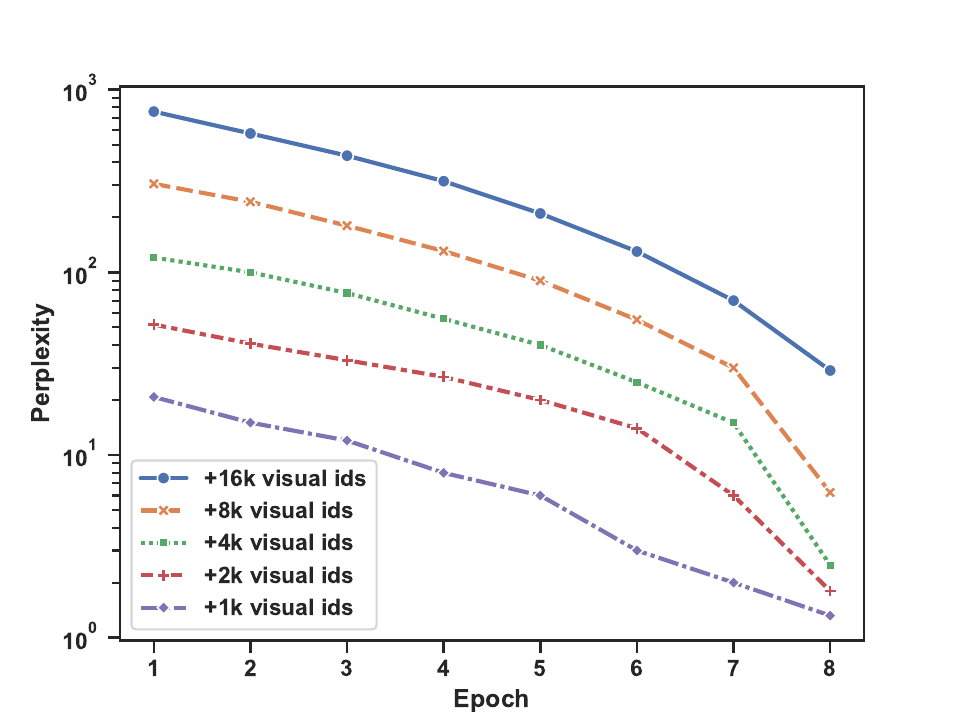}
\caption{The perplexity trajectories of vanilla unified autoregressive models with different sized visual vocabularies.}
\label{ppl}
\end{figure}

Current vanilla unified autoregressive works \cite{team2024chameleon, wang2024emu3, wu2024liquid} directly mix data from different modalities for joint learning. In particular,  all the visual token IDs are activated and directly integrated into the training process. However, we suspect it may affect the fusion process across modalities and ultimately lead to suboptimal performance on various tasks. To further illustrate this issue, we conducted an intuitive experiment. As shown in Figure \ref{ppl}, we list the perplexity trajectories of vanilla unified autoregressive models trained with different sized visual vocabularies. It shows that as the size of visual vocabulary gets larger, the corresponding perplexity score of the unified model increases significantly. These findings have validated our previous suspect and provided valuable insights, which informed the development of the proposed progressive vocabulary learning method. Thus, the newly proposed progressive vocabulary learning method is illustrated as follows, which is conducted during the \textbf{Unified Pretraining stage of UGen}.


\begin{itemize}[label=\textbullet, leftmargin=*]
    \item Firstly, UGen is trained to acquire strong text processing capabilities with text-only data. 
    \item Second, a mix of text-only, image-to-text and text-to-image data is used to train UGen to obtain the capabilities of image understanding and generation. Instead of directly incorporating entire visual vocabulary for training, the visual IDs are gradually activated and joint learned with the existing textual token IDs progressively. The activation strategy of the visual token IDs is simple but effective, which is detailed in algorithm \ref{algo}. Specifically, for every $k$ training step, we randomly select one visual token id $v$ from the visual vocabulary and then the selected id is integrated into the training phase. For an input multimodal id sequence (described in Section \ref{Architecture}), the activated visual token IDs remain unchanged, while the unactivated visual token IDs are replaced with a special token $[MASK]$. 
    \item After all the visual IDs are integrated into the training phase, the pretraining process continues until the unified model converges.
\end{itemize}

\noindent In our experiments, we test different $k$ values to achieve optimal performance, which will be shown in the experimental section.

\begin{algorithm}[t]
\small
\begin{algorithmic}
\caption{The \textbf{activation strategy} for visual token IDs in progressive vocabulary learning}
\label{algo}
\STATE \textbf{Input:} Training Dataset $D$; UGen model $f(\cdot|\theta)$; Textual, visual and activated Vocabulary $V_T$, $V_I$, $V_A$; Hyperparamters step $k$. Training steps $t$.
\STATE \textbf{Initialize:} $V_A$ = $V_{T}; t=0;$ 
\newline
\WHILE{Training}
    \STATE $t=t+1;$
    \STATE \textcolor{lightgray}{\texttt{//} Randomly select one visual token id for every k steps}
    \vspace{-1em}
    \IF{$t$ $\%$ $k == 0$ and $V_I \ne \oslash $} 
        \STATE $v_t=random.choice(V_I)$ 
        \STATE $V_A.add(v_t)$ \textcolor{lightgray}{\texttt{//} Active a visual token ID}
        \STATE $V_I.del(v_t)$ \textcolor{lightgray}{\texttt{//} Delete the selected visual token ID}
    \ENDIF

    \STATE \textcolor{lightgray}{\texttt{//} Randomly select a sample $x$ from the training set D}
    \STATE $x = (x_1,x_2,...,x_n) = random.choice(D)$
    \STATE \textcolor{lightgray}{\texttt{//}  Unactivated visual tokens are masked}
    \FOR{$x_i$ in  $(x_1, x_2, ..., x_n)$}
        \IF{$x_i$ not in $V_A$}
            \STATE $x_i = [MASK]$
        \ENDIF
    \ENDFOR
    \STATE $loss=f(x|\theta)$
    \STATE Backward $loss$ and update $\theta$
\ENDWHILE
\end{algorithmic}
\end{algorithm}


\subsection{Training and Inference Procedure}
\label{Training and Inference Procedure of UGen}

\paragraph{Training} we adopt the cross-entropy loss during training as below: 
\begin{equation}
\mathcal L=-\sum_{i=1}^{}logP_{\theta}(x_i|x_{<i})
\end{equation}
    
\noindent Here, $P(\cdot |\cdot )$ denotes the conditional probability modeled by the weight $\theta$ of UGen. For all language-vision understanding and generation tasks, we compute the loss on the whole sequence. To keep the design simple, all the multimodal tasks share the same loss weights. Overall, the training paradigm of UGen is divided into two stages. Details are provided below.

\begin{itemize}[label=\textbullet, leftmargin=*]
    \item \textbf{Unified Pretraining} (Pretrain): In this stage, we perform unified pretraining to enable UGen to acquire the capabilities of text processing, image understanding and image generation. In addition, all the parameters of UGen is activated and trained on a mix of multimodal data, including text-only, image-to-text, and text-to-image data. The ratio of different modality data is illustrated in the experimental section. It is critical to note that \textbf{we employ the progressive vocabulary learning approach in this stage}, which differs from previous works. 

    \item \textbf{Supervised Fine-tuning} (SFT): We further fine-tune the pretrained model with instruction tuning data to enhance its instruction-following capability. All the parameters are tuned. Moreover, unlike other works \cite{wang2024emu3, wu2024janus, wu2024liquid} which fine-tune separated models, we train a single model using a blend of text-only, image-to-text and text-to-image data to ensure versatility across various scenarios.
\end{itemize}

\paragraph{Inference} UGen adopts a next-token prediction approach. For text processing and image understanding tasks, we follow the standard practice of sampling tokens sequentially from the predicted distribution. For the image generation task, we utilize classifier-free guidance (CFG),  similar to previous works \cite{chang2023muse, gafni2022make, sun2024autoregressive}. Specifically, for each token, the logit $l_g$ is calculated as: 

\begin{equation}
    l_g=l_u+s(l_c-l_u)
\end{equation}

\noindent $l_c$ is the conditional logits, $l_u$ is the unconditional logits, and $s$ is the scale for the classifier-free guidance. The default number of $s$ is 5.

\section{Experiments}

\begin{table}[t]
    \centering
    \begin{tabular}{c|ccc}
        \toprule
        \textbf{Hyperparameters} & \textbf{Pretrain} &  \textbf{SFT} \\
        \hline
        Learning rate & $1.0\times 10^{-4}$ & $3.0\times 10^{-5}$ \\
        LR scheduler & Cosine & Cosine \\
        Weight decay & 0.01 & 0.01\\
        Gradient clip & 1.0 & 1.0\\
        Optimizer & AdamW & AdamW \\
        Warm-up steps & 5,000 & 1,000 \\
        Training steps & 500,000 & 10,000\\
        Batch size & 256 & 128\\
        Data Ratio & 3:2:5 & 2:6:2 \\
        \bottomrule
    \end{tabular}
    \caption{Detailed hyperparameters of the Pretain and SFT stage of UGen. Data ratio refers to the ratio of text-only, image-to-text, and text-to-image data.}
    \label{hyper}
\end{table}

\subsection{Implementation Details}
\label{implementation-details}

As mentioned above, UGen is first pretrained to acquire strong text processing capabilities. In our experiments, for minimizing the training costs, we initialized UGen with a pre-trained TinyLlama(1.1B) \cite{zhang2024tinyllama} model, which has a 32k sized textual vocabulary and supports a maximum sequence length of 4,096. Moreover, we adopt SBER-MoVQGAN-67M\footnote{https://github.com/ai-forever/MoVQGAN} to quantize images into discrete tokens, which has a codebook size of 16,384 and achieves $8\times8$ compression in the spatial dimension. All images are resized to 256 $\times$ 256 pixels. Above all, the detailed hyperparameters for each training stage are provided in Table \ref{hyper}. We use sequence packing during training to improve training efficiency. We mix all the multimodal data according to the specified ratios in a single training step. 
 

\subsection{Pretrain and SFT Datasets}
\label{data-setup}
In this section, we provide details of the Pretrain and SFT datasets. 


\paragraph{Pretrain Datasets} 

The pretraining data consists of three components. (i) Text-only data, which consists of 30M text instances for approximately 30B textual tokens. Specifically, we randomly sampled 15M data from DCLM \cite{li2024datacomp} and another 15M data from SlimPajama \cite{soboleva2023slimpajama} to construct the whole dataset. (ii) Image-to-text data, which contains 30M image-text pairs: 1M samples from ImageNet \cite{deng2009imagenet}, 4M samples from JourneyDB \cite{sun2024journeydb}, 12M samples from CC12M \cite{changpinyo2021conceptual}, 12M samples from LAION-aesthetics-12M \footnote{https:\//\//huggingface.co\//datasets\//dclure\//laion-aesthetics-12m-umap} and 1M in-house data. We adopt the open-source CogVLM \cite{wang2023cogvlm} model to re-caption images for all the image-to-text data. (iii) Text-to-image data, which is the same as image-to-text data, except that each image and text pair is reversed. In addition, as CFG strategy \cite{chang2023muse, gafni2022make, sun2024autoregressive} is used in image generation task, 10\% of the image captions are randomly dropped. 

\begin{table*}[t]
    \centering
    \begin{tabular}{cccccc}
        \toprule
         \textbf{Method} & \textbf{Type} & \textbf{\# Params} & \textbf{Text} & \textbf{Image-Und} & \textbf{Image-Gen}  \\
         \hline
    TinyLlama          & Text Only      & 1.1B & 52.9 & -    & - \\
    LLaVA-v1.5-Phi-1.5 & Image-Und Only & 1.3B & -    & 68.1 & - \\
    MobileVLM          & Image-Und Only & 1.7B & -    & 60.4 & - \\
    LLaVA-v1.5         & Image-Und Only & 7B   & -    & 75.4 & - \\
    mPLUG-Owl2         & Image-Und Only & 7B   & -    & 73.4 & - \\
    SDv2.1             & Image-Gen Only & 0.9B & -    & -    & 50.0 \\
    SDXL               & Image-Gen Only & 2.6B & -    & -    & 55.0 \\
    Jauns              & Image-Und\&Gen & 1.5B   & 41.0\textsuperscript{*} & 72.5 & 61.0 \\
    Show-o             & Image-Und\&Gen & 1.3B   & 36.1\textsuperscript{*} & 57.3 & 53.0 \\
    Chameleon          & All & 7B     & 68.4                    & -    & 39.0 \\
         \midrule[0.8pt]
    Task-specific AR   & Text Only      & 1.1B & 48.8 & -    & - \\
    Task-specific AR   & Image-Und Only & 1.1B & -    & 66.4 & - \\
    Task-specific AR   & Image-Gen Only & 1.1B & -    & -    & 55.9 \\
    Vanilla Unified AR & All & 1.1B & 42.1  & 61.0  & 42.5  \\
        \hline
    \textbf{UGen} & All & 1.1B & 46.1 & 66.0 & 52.1 \\
        \bottomrule
    \end{tabular}
    \caption{The main experimental results of UGen, which consists of two main parts. First, a strictly comparison (lines 11-15) is conducted between UGen and vanilla AR models, which are trained under single-task specific modeling (Task-specific AR) and multi-task unified modeling (Vanilla Unified AR) conditions respectively. Besides, all the models are trained with the same data, model backbone and training configurations. Second, UGen is compared with other mainstream models on various tasks (Lines 1-10, and 15).}


    \label{table-main}
\end{table*}

\paragraph{SFT Datasets} For text processing, we collected total 118K data from \citet{clark2019boolq, bisk2020piqa, zellers2019hellaswag, sakaguchi2021winogrande, clark2018think, mihaylov2018can} and the additional 1M in-house data. For image understanding, we collected total 1.2M instruct tuning data from \citet{goyal2017making, hsiao2022screenqa, hudson2019gqa, li2024llava, lu2021iconqa, shah2019kvqa, zhu2024llava} and 1M in-house data. For image generation, we collected total 4.4M image-to-text pairs from \citet{sun2024journeydb, singla2024pixels} and 1M in-house data. We utilize the following formatted instruction: "User:<Input Message> /\/n Assistant:<Response>". For multi-turn dialogues, we repeat this format to structure the data.

\subsection{Evaluation Setup}
\label{evaluation-setup}
We conduct evaluations on comprehensive benchmarks, which are listed as follows.

\paragraph{Text Processing Tasks} To assess the capabilities of text processing, we evaluate UGen on widely recognized benchmarks, which include HellaSwag \cite{zellers2019hellaswag}, WinoGrande \cite{sakaguchi2021winogrande}, ARC-Easy \cite{clark2018think}, ARC-Challenge \cite{clark2018think}, OpenBookQA \cite{mihaylov2018can}, PIQA \cite{bisk2020piqa}, SIQA \cite{sap2019socialiqa} and BoolQ \cite{clark2019boolq}. We adopt the macro-average score of the 8 datasets as the overall performance score of textual processing capabilities.

\paragraph{Image Understanding Tasks} We evaluate UGen on a various of image understanding benchmarks, including VQAv2 \cite{goyal2017making}, GQA\cite{hudson2019gqa}, MME\cite{fu2023mme}, and POPE\cite{li2023evaluating}. The macro-average score of the 4 datasets is employed as the final metrics for image understanding capabilities. 

\paragraph{Image Generation Tasks} GenEval \cite{ghosh2024geneval} is used to evaluate UGen's image generation capabilities. GenEval is a challenging image generation benchmark, which is designed to evaluate the comprehensive generative abilities of image generation by offering a detailed instance-level analysis of their compositional capabilities.

\subsection{Main Results}
\label{main-results}
Two types of experiments are designed in this section (table \ref{table-main}). First, we design a set of strictly comparable experiments to illustrate the challenges faced by vanilla unified AR modeling and the effectiveness of UGen (lines 11-15). Second, we further compared the performance of UGen with some mainstream models across various of tasks, in order to demonstrate the reliability of the experiments (lines 1-10 and 15). Moreover, to the best of our knowledge, we are the first to conduct a rigorous comparison of autoregressive models under multi-task unified modeling and single-task specific modeling conditions, using the same data, model backbone, and training configurations.

According to the experiment results, it is evident that vanilla unified AR model suffers from inferior performance compared to task-specific AR models. The magnitude of the performance decline varies across different tasks, ranging from 8.1\% to 23.9\%. In contract, UGen achieves strong performance on all the language-vision understanding and generation tasks. Compared with vanilla unified AR model, UGen achieves a significant improvement of 13.3\% on tested tasks. Compared with the task-specific AR models, UGen significantly reduces the magnitude of performance decline to 0.6\% and 7.3\%. The above experimental results fully demonstrate the effectiveness of the novel progressive visual vocabulary learning method in UGen. It also suggests that by reducing the mutual interference among various tasks during joint training, it helps to achieve excellent performance across all tasks.


\begin{figure*}[t]
\centering
\includegraphics[width=1\textwidth]{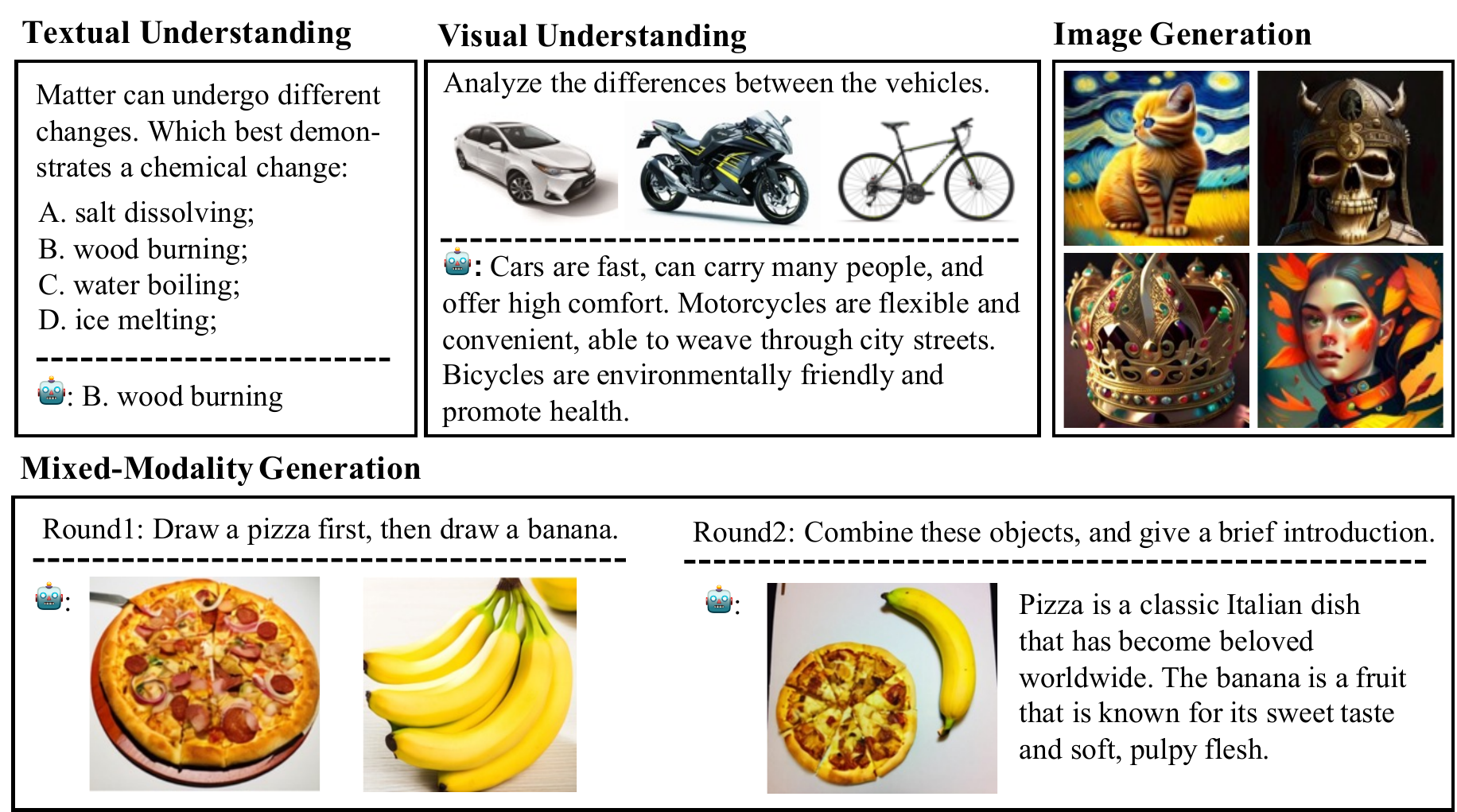}
\caption{\textbf{The Qualitative results of UGen}}
\label{case}
\end{figure*}




Furthermore, we find that UGen achieves comparable performance to mainstream models across various tasks. On text processing task, it achieves consistent results with LLaVA-v1.5-Phi-1.5 and MobileVLM, which are specifically designed. On task of image generation, UGen surpasses SDv2.1 and gets comparable results with SDXL, which suggests that it possesses strong image generation capability. When compared with other unified multimodal models, UGen also surpasses Show-o and Chameleon, which are well-known models of this field. Moreover,  Chameleon has 7B parameters, far exceeding the parameter scale of UGen. The experimental results above strongly demonstrate that UGen has achieved highly competitive performance across various tasks. And it further demonstrates the high credibility of the experiments in this paper. However, we also acknowledge that the current performance of UGen is inferior to the SOTA model, Janus. We believe that the main reasons lie in the scale of the training data and the limitations of computing resources, which will be the focus of our future work.

\subsection{Detailed Analysis}
\label{detailed-analysis}
We give various detailed analysis in this section and demonstrate the following conclusions. (i) UGen is versatile and masters a novel capability of mixed-modality generation. (ii) UGen achieves a stable learning curve compared with vanilla unified AR model. (iii) The progressive vocabulary learning is robust and consistently improves the performance. (iv) UGen performs well even when it is training from scratch. 

\paragraph{Qualitative results of UGen} Figure \ref{case} shows the diverse results generated by UGen to illustrate its versatility. It demonstrates that UGen can produce high-quality images and text. Interestingly, thanks to the unified modeling, UGen masters a novel capability of mixed-modality generation, and is able to generate interleaved textual and image content.



\begin{figure}[t]
\centering
 \includegraphics[width=0.96\linewidth]{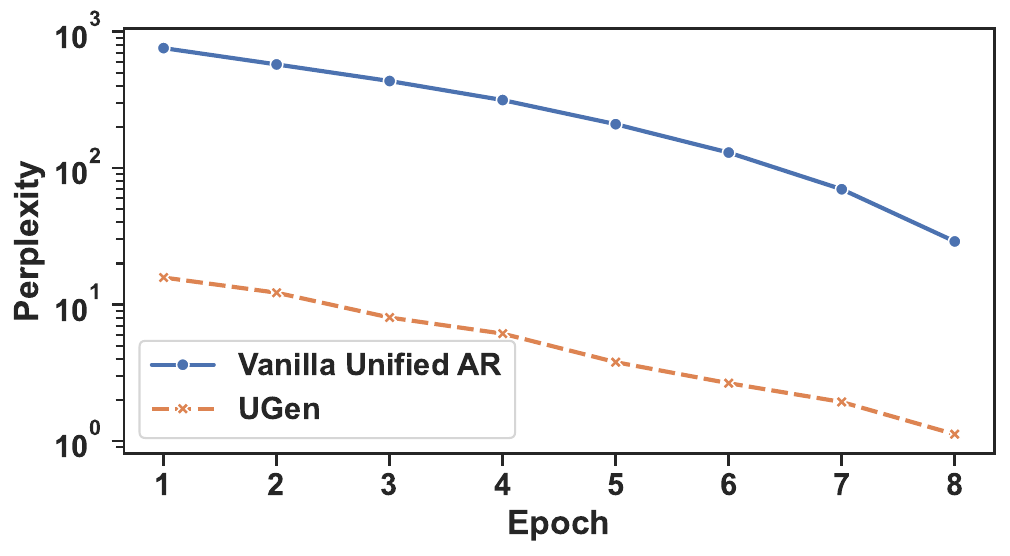}
\caption{The perplexity trajectories of UGen and vanilla unified AR model. The horizontal axis represents the training epochs of the models. The vertical axis denotes the perplexity score of the generative model.} 
\label{ppl_our}
\end{figure}

\paragraph{Perplexity trajectories of UGen} As mentioned in section \ref{Progressive Vocabulary Learning}, the vanilla unified AR training approach leads to a significant increase in the model's perplexity scores, which indicates the exist of mutual interference between textual and visual modalities. Figure \ref{ppl_our} shows the perplexity trajectories of UGen and vanilla unified AR model. We observe that the perplexity score of UGen is maintained at a relatively low level and UHen achieves a relatively stable learning curve. It further proves that our proposed  vocabulary learning method can effectively reduce the mutual interference between textual and visual tasks, which helps UGen to achieve strong results across various tasks.

\paragraph{Impact of visual token IDs' activation speeds} 

\begin{figure}[t]
\centering
\includegraphics[width=1.02\linewidth]{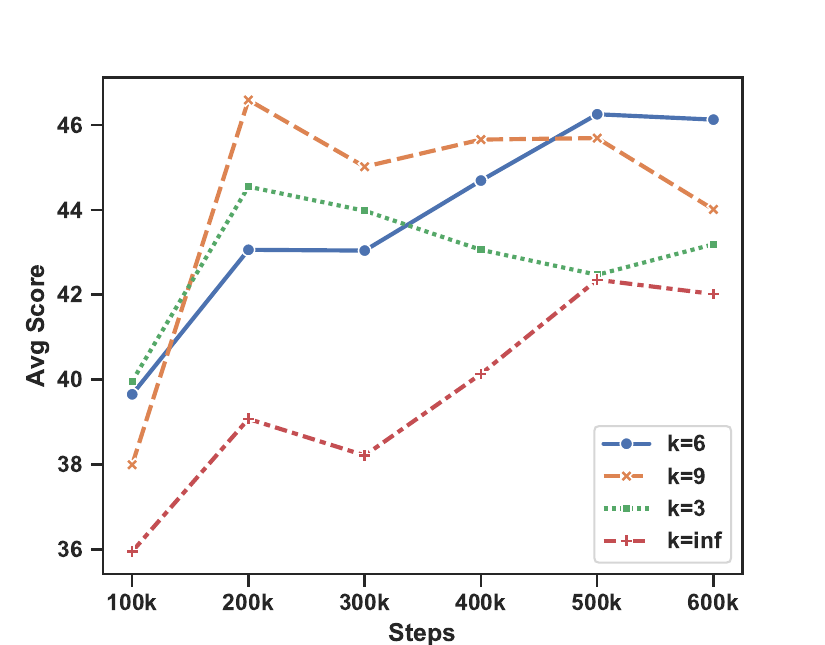}
\caption{Impact of visual token IDs activation speed. The horizontal axis represents the training steps. The vertical axis denotes the average performance score of all tasks. When we set $k=inf$, UGen is equivalent to the vanilla unified AR model.} 
\label{speed}
\end{figure}

As mentioned in section \ref{Progressive Vocabulary Learning}, we conduct additional experiments to examine how different activation speeds affect the model performance, which is shown in figure \ref{speed}. We conclude that our approach is robust and can consistently improve the performance at different activation speeds. Furthermore, excessively fast or slow activation speeds lead to suboptimal performance, whereas an appropriate activation speed achieves the best results.



\begin{figure}[t]
\centering
\includegraphics[width=1.02\linewidth]{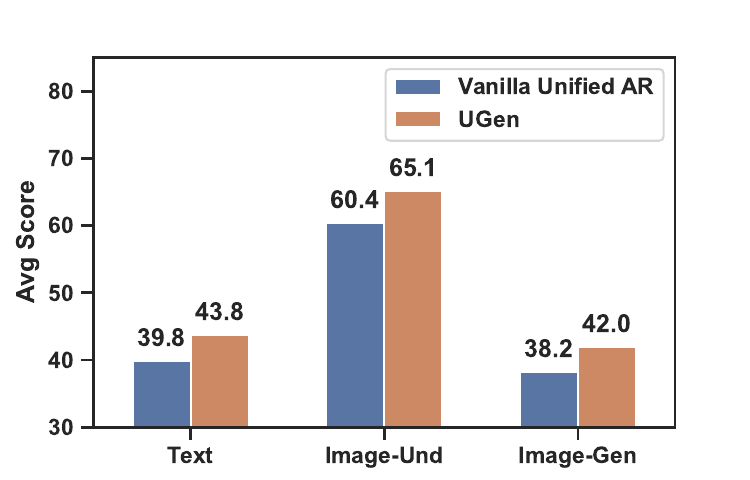}
\caption{The performance comparison of UGen and vanilla unified AR model trained from scratch. The horizontal axis represents different tasks. The vertical axis denotes the performance scores.}   
\label{cold}
\end{figure}

\paragraph{Training from scratch for UGen}
Another experiment is conducted to test the performance of UGen trained from scratch. Therefore, we randomly initialize all model parameters, then train UGen and vanilla unified AR model with the same data and settings. From figure \ref{cold}, we conclude that UGen also perform well on various tasks when training from scratch, when it is compared with the vanilla unified AR model. This further demonstrates that the effectiveness of UGen is not limited to the warm-started conditions.


\section{Related Work}

Recently, there have been many efforts to use a single transformer to uniformly model the language-vision understanding and generation tasks. LWM \cite{liu2024world}, Chameleon \cite{team2024chameleon} and AnyGPT \cite{zhan2024anygpt} are pioneering works, which quantize images into discrete tokens and are trained on a mix of multimodal data. Their autoregressive architecture allows for seamless information integration across various modalities. However, in practice, these works fail to achieve high performance on all multimodal tasks concurrently. 

Several works proposed to alleviate the problem by introducing additional compositions. Firstly, some works focused on the approach for multimodal data encoding. For example, \cite{wu2024janus, chen2025janus} proposed to decouple visual encoding for multimodal understanding and generation, which could reduce the conflict arising from the differing demands that understanding and generation place on the visual encoder. Secondly, from the perspective of model architecture, \cite{zhou2024transfusion, zhao2024monoformer, xie2024show} proposed to combine the language modeling loss function
(next token prediction) with diffusion to train a single transformer, in order to bridge the gap between the discrete sequence modeling and continuous image generation. Another work \cite{shi2024llamafusion} proposed to freeze the initial LLM and integrate additional components for image tasks, which is designed to reduce mutual influence across language-vision modalities. Last but not the least, from the perspective of training procedure, \cite{wang2024emu3} provided a more direct solution, to finetune task-specific models for each task. Although these methods differ, the additional compositions and parameters significantly increase the model complexity. 

In this paper, we pursue to achieve promising performance across various tasks without any additional components, while maintaining the simple autoregressive architecture. A contemporaneous work \cite{wu2024liquid} also focused on this issue and proposed to achieve promising results by scaling the model parameters. In contract, we explored to achieve strong performance results under the same model scale and data constraints. We are also interested to analyze the impact of data and model scales for the unified autoregressive modeling in the future.



\section{Conclusion and Future Works}
In this paper, we study the unified modeling for language-vision understanding and generation. We propose UGen, which is a unified autoregressive multimodal model and trained with a novel progressive vocabulary learning method. Experiments on various multimodal benchmarks show that, UGen achieves competitive results with task-specific models and also yields novel capabilities of mixed-modality generation. To the best of our knowledge, this is the first work to boost both textual and visual capabilities with a simple autoregressive architecture. In the future, we hope to enhance UGen by scaling up the data, models, and modalities.


\label{sec:bibtex}

\bibliography{custom}

\appendix

\section{Example Appendix}
\label{sec:appendix}

\begin{figure*}[!ht]
\centering
\includegraphics[width=1\textwidth]{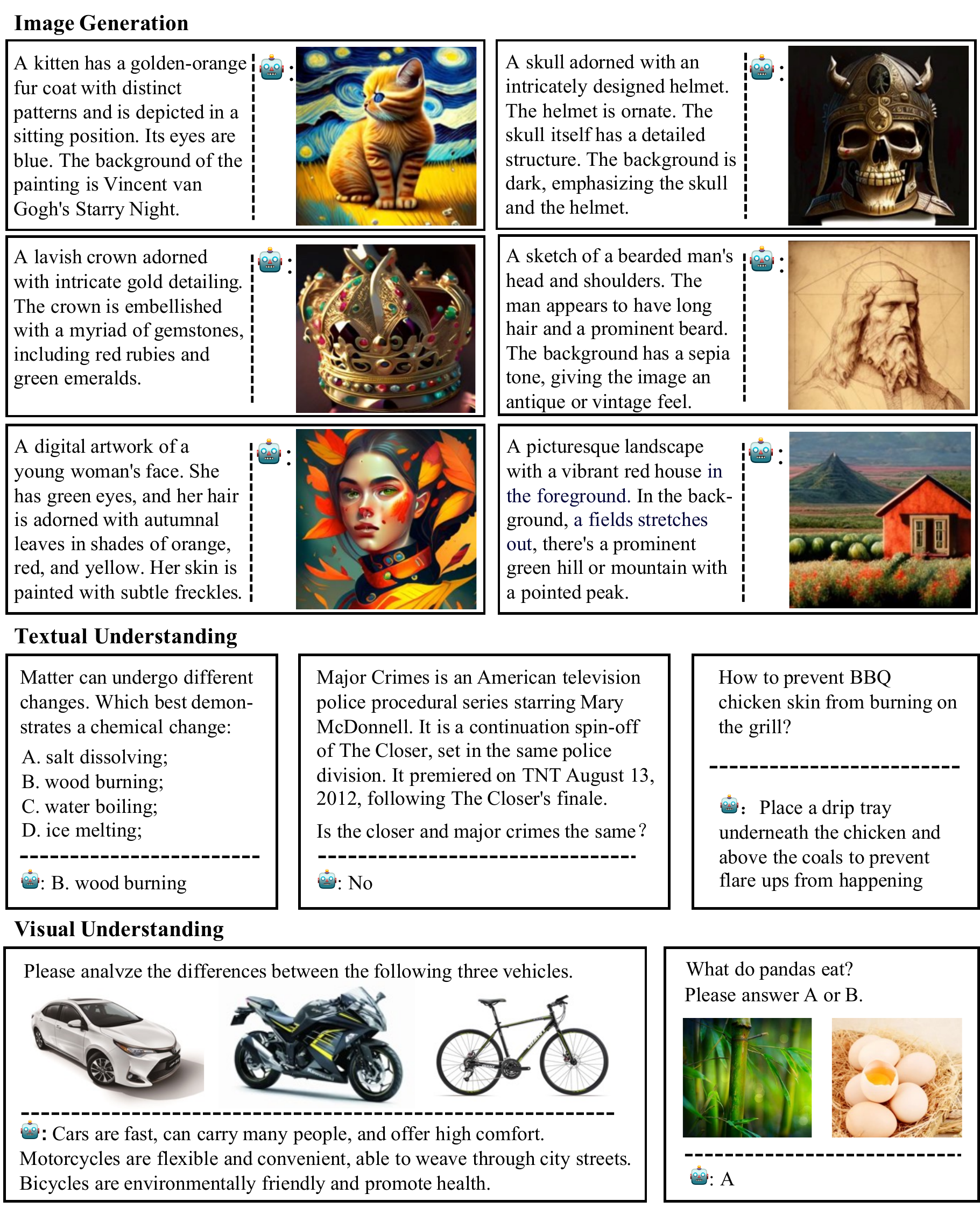}
\caption{\textbf{The Qualitative results of UGen.}}
\label{more-case}
\end{figure*}

The performance of UGen, compared among vanilla unified autoregressive model (denotes as Vanilla Unified AR), task-specific autoregressive models (denotes as Task-specific AR), and previous work models on various tasks are shown in the following tables. More qualitative results of UGen are listed in Figure \ref{more-case}. 
\newpage

\begin{table*}[t]
    \centering
    \begin{tabular}{ccccccccccc}
        \hline
         Method & \# Params & BoolQ & PIQA& HS& WG& OBQA& ARC-e& ARC-c & Avg. & \\
        \hline
        TinyLlama & 1.1B & 57.8 & 73.3 & 59.2 & 59.1 & 36.0 & 55.2 & 30.1 & 52.9 \\
        Janus\textsuperscript{*} & 1.5B & 46.3 & 58.4 & 28.7 & 55.2 & 37.6 & 29.8 & 29.1 & 40.7 \\
        Show-o\textsuperscript{*} & 1.5B & 42.2 & 48.3 & 23.9 & 48.5 & 36.4 & 26.0 & 27.3 & 36.0 \\
        Chameleon & 7B & 81.4 & 79.6 & 74.2 & 70.4 & 51.0 & 76.1 & 46.5 & 68.4 \\
        \hline
        Task-specific AR & 1.1B & 55.1 & 72.4 & 44.9 & 51.4 & 27.2 & 60.4 & 30.5 & 48.8 \\
        Vanilla Unified AR & 1.1B & 55.9 & 64.9 & 35.7 & 48.8 & 25.0 & 37.2 & 27.1 & 42.1 \\
        \hline
        \textbf{UGen (Our)} & 1.1B & 56.7 & 70.3 & 46.3 & 50.4 & 26.4 & 43.8 & 28.7 & \textbf{46.1} \\
        \hline
    \end{tabular}
    \caption{The performance of UGen, compared among vanilla unified autoregressive model (denotes as Vanilla Unified AR), task-specific autoregressive models (denotes as Task-specific AR), and previous work models on text processing task. WG is short for WinoGrande, HS is short for HellaSwag. * indicates the score is evaluated by released model.}
    \label{tablename2}
\end{table*}

\begin{table*}[t]
    \centering
    \begin{tabular}{ccccccc}
        \hline
         Method & \# Params & VQAv2 & GQA & MME (Norm.) & POPE & Avg. \\
        \hline
        LLaVA-v1.5-Phi-1.5 & 1.3B & 75.3 & 56.5 & 1128.0 (56.4) & 84.1 & 68.1 \\
        MobileVLM & 1.7B & 56.1 & 41.5 & 1196.2 (59.8) & 84.5 & 60.4  \\
        LLaVA-v1.5 & 7B & 85.9 & 62.0 & 1510.7 (75.5) & 78.5 & 75.4  \\
        mPLUG-Owl2 & 7B & 79.4 & 56.1 & 1450.2 (72.5) & 85.8 & 73.4  \\
        Janus & 1.5B  & 77.3 & 59.1 & 1338.0 (66.9) & 87.0 & 72.5  \\
        Show-o & 1.3B & 59.3 & 48.7 & 948.4 (47.4) & 73.8 & 57.3  \\
        \hline
        Task-specific AR & 1.1B & 73.1 & 54.9 & 1128.0 (56.4) & 81.2 & 66.4 \\
        Vanilla Unified AR & 1.1B & 65.0 & 52.1 & 1065.0 (53.2) & 73.9 & 61.0 \\
        \hline
        \textbf{UGen (Our)} & 1.1B & 70.0 & 58.2 & 1110.0 (55.5) & 80.4 & \textbf{66.0} \\
        \hline
    \end{tabular}
    \caption{The performance of UGen, compared among vanilla unified autoregressive model (denotes as Vanilla Unified AR), task-specific autoregressive models (denotes as Task-specific AR), and previous work models on image understanding task. }
    \label{tablename3}
\end{table*}

\begin{table*}[t]
    \centering
    \begin{tabular}{ccccccccc}
        \hline
         Method & \# Params & \makecell{Single \\ Obj.} & \makecell{Two \\ Obj.} & Counting& Colors & Position & \makecell{Color \\ Attri.} & Overall \\
        \hline
        SDv2.1   & 0.9B & 98.0 & 51.0 & 44.0 & 85.0 & 7.0  & 17.0 & 50.0 \\
        SDXL     & 2.6B & 98.0 & 74.0 & 39.0 & 85.0 & 15.0 & 23.0 & 55.0 \\
        Janus    & 1.5B & 97.0 & 68.0 & 30.0 & 84.0 & 46.0 & 42.0 & 61.0 \\
        Show-o   & 1.3B & 95.0 & 52.0 & 49.0 & 82.0 & 11.0 & 28.0 & 53.0  \\
        Emu3-Gen & 8B   & 98.0 & 71.0 & 34.0 & 81.0 & 17.0 & 21.0 & 54.0  \\
        \hline
        Task-specific AR & 1.1B & 88.7 & 58.6 & 45.0 & 75.5 & 35.0 & 32.0 & 55.9 \\
        Vanilla Unified AR & 1.1B & 87.5 & 43.4 & 17.5 & 72.3 & 25.0 & 10.0 & 42.5 \\
        \hline
        \textbf{UGen (Our)} & 1.1B & 91.2 & 56.7 & 33.7 & 73.4 & 26.0 & 36.0 & \textbf{52.1} \\
        \hline
    \end{tabular}
    \caption{The performance of UGen, compared among vanilla unified autoregressive model (denotes as Vanilla Unified AR), task-specific autoregressive models (denotes as Task-specific AR), and previous work models on image generation task. }
    \label{tablename4}
\end{table*}

\end{document}